# Impact of Data Pruning on Machine Learning Algorithm Performance

Debrup Chakraborty
School of Computer Science & Statistics
Trinity College Dublin
Dublin, Ireland
chakrabd@tcd.ie

Viren Chhabria
School of Computer Science & Statistics
Trinity College Dublin
Dublin, Ireland
chhabriv@tcd.ie

Aneek Barman Roy
School of Computer Science & Statistics
Trinity College Dublin
Dublin, Ireland
barmanra@tcd.ie

Arun Thundyill Saseendran
School of Computer Science & Statistics
Trinity College Dublin
Dublin, Ireland
thundyia@tcd.ie

Lovish Setia
School of Computer Science & Statistics
Trinity College Dublin
Dublin, Ireland
setial@tcd.ie

*Abstract:* Dataset pruning is the process of removing sub-optimal tuples from a dataset to improve the learning of a machine learning model. In this paper, we compared the performance of different algorithms, first on an unpruned dataset and then on an iteratively pruned dataset. The goal was to understand whether an algorithm (say A) on an unpruned dataset performs better than another algorithm (say B), will algorithm B perform better on the pruned data or vice-versa. The dataset chosen for our analysis is a subset of the largest movie ratings database publicly available on the internet, IMDb [1]. The learning objective of the model was to predict the categorical rating of a movie among 5 bins: poor, average, good, very good, excellent. The results indicated that an algorithm that performed better on an unpruned dataset also performed better on a pruned dataset.

## 1 INTRODUCTION

A fine line separates cleaning and pruning of a dataset. Cleaning mostly is a preprocessing step that involves removing unrequired data, data imputation, standardizing or normalizing the feature ranges and converting categorical values to numbers [2] [3]. In comparison pruning takes place after preprocessing, where certain data is strategically removed to improve the machine learning model. In this paper we try to bring forth the effect of dataset pruning on the performance of different machine learning algorithms, i.e. If an algorithm (say A) on an unpruned dataset performs better than another algorithm (say B), will algorithm B perform better on the pruned data or vice-versa.

## 2 RELATED WORK

Data pruning had been defined in 2005 as an automated process of noise cleaning and the performance of this mechanism was measured using SVC and AdaBoost algorithms [4]. Removal of certain portions of the dataset is determined to be worthwhile and said to affect the performance of machine learning algorithms [4]. A mathematical model was proposed to predict the success of upcoming movies based on correlation of factors affecting the success of a movie [5].

Automatic rating prediction was proposed in 2011 using the IMDb dataset, however the results were inferior to baseline which was attributed to the dataset lacking diversity in terms of user rating [6].

## 3 METHODOLOGY

### 3.1 Dataset

The dataset chosen is from the largest publicly available movie rating database, IMDb [1]. It contains 5,043 movies with 28 attributes, with IMDb score indicating the movie ratings on a scale of 1-10. The histogram in Figure 1 shows the frequency distribution of the IMDb score indicating the rating between 6 and 7 to be the highest.

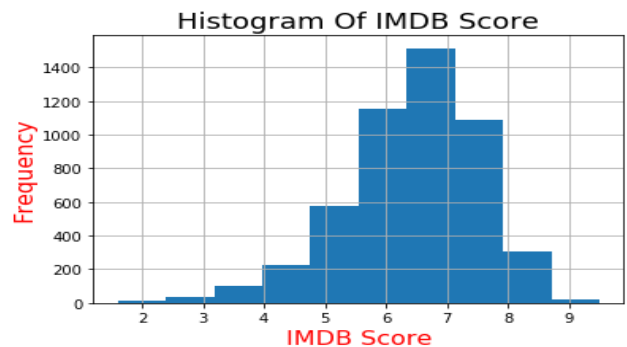

**Figure 1: Frequency of IMDb Score of raw dataset**

### 3.2 Pre-processing

IMDb ratings have continuous values in the range 1-10. The ratings were categorized into 5 classes: poor, average, good, very good, excellent based on the bins [0, 7, 8, 8.5, 9, 10]. Missing numeric data was imputed with the mean of the available values, while the missing categorical data was imputed as a "Missing" category altogether. Duplicate tuples were removed. Categorical data was transformed to numbers using LabelEncoder and OneHotEncoder. The feature data was standardized using StandardScaler. The aforementioned utilities were used from the scikit-learn library [7] and the code base is available in an open source code repository [11]. Figure 2 represents the correlation between the features used for training the model.

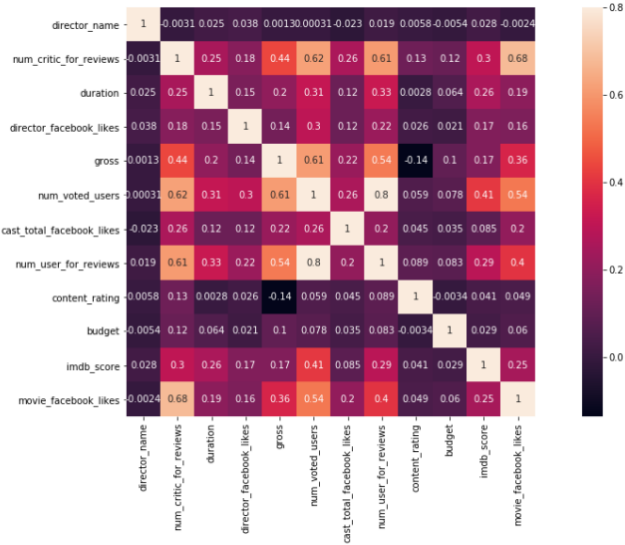

**Figure 2: Correlation between the selected features**

## 3.3 Data Pruning

The pre-processed dataset was then pruned based on the number of user reviews for a movie. We iteratively pruned the dataset where a movie had received less than {1...20} user reviews. This was done as a lower review count would make the rating of that movie biased to a small (<20) number of user opinions. Figure 3 shows a scatter plot depicting the number of user reviews v/s IMDb score.

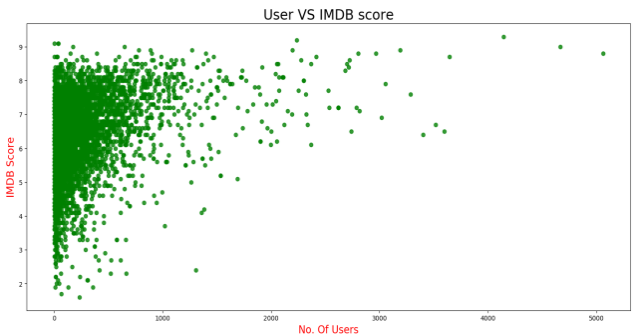

**Figure 3: Plot of IMDB score v/s number of users who reviewed on a raw dataset**

## 3.4 Algorithms

We used Logistic Regression, Random Forest Classifier and Support Vector Classifier (SVC) to evaluate the sensitivity of machine learning algorithms to data pruning. Parameter tuning for these algorithms was done while training the model using the unpruned dataset, and the best measure from our findings was used for each iteration of the pruned dataset.

*3.4.1 Logistic Regression:* We used different values for the algorithm parameter c (inverse of the regularization strength) in the range 0.001 to 1000 with 10x increments. We noticed the highest accuracy for c=10 as shown in Figure 4.

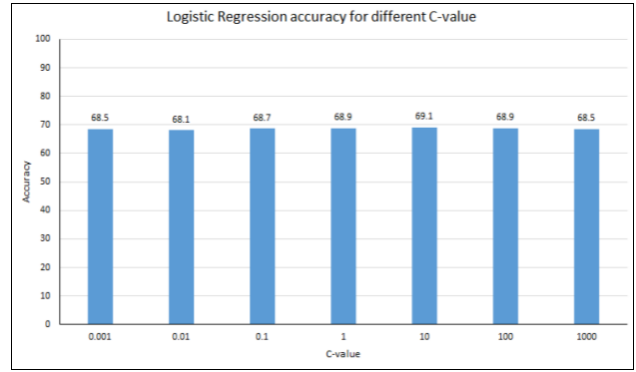

**Figure 4: c-value accuracy of logistic regression**

*3.4.2 Random Forest:* The n-estimator (number of decision tree classifiers) for random forest was experimented with in the range 10 to 100 in increments of 10, and it was found to be best at 40 as shown in Figure 5.

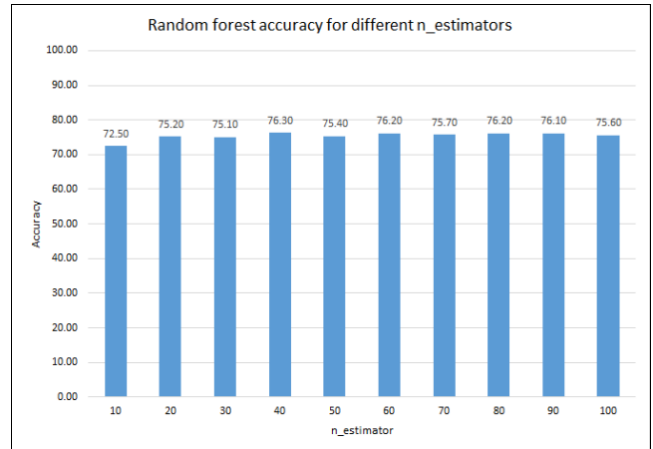

**Figure 5: n-Estimator accuracy for random forest**

*3.4.3 Support Vector Machine (SVM):* We ran the SVM weights to prevent overfitting on larger margins. For c (regularization parameter) in the range [0.001, 0.01, 0.1, 10, 25, 50, 1000], SVM was found to perform best for c =25 as shown in Figure 6.

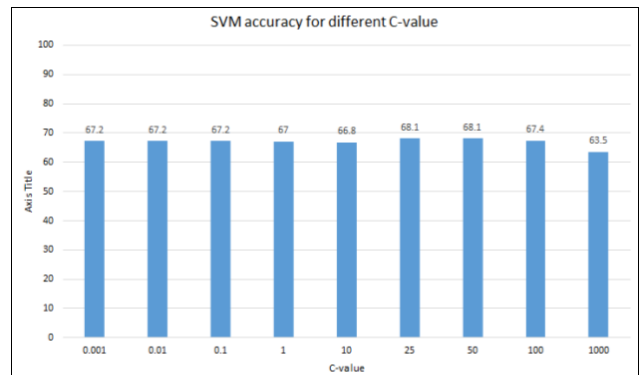

**Figure 6: c-value accuracy of SVM**

## 3.5 Evaluation

We tried splitting the dataset into train-test sets in 3 different ratios and found that 20% test data would give an able accuracy measure. This was manually tested and eventually we chose the 80:20 split ratio for train-test datasets as shown in Figure 7.

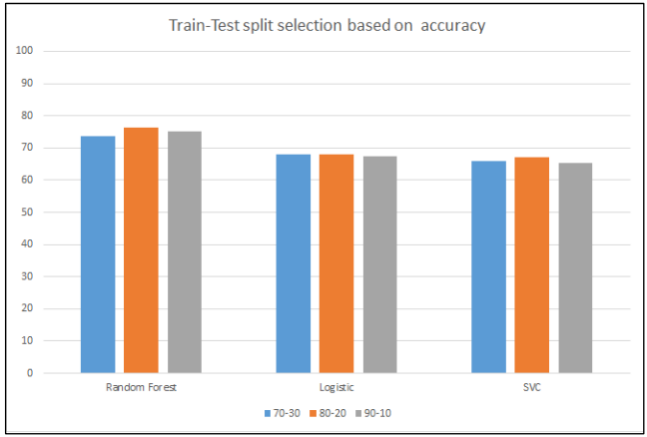

**Figure 7: Accuracy of train-test dataset split for various algorithms**

## 4 RESULTS AND DISCUSSION

### 4.1 Metrics

The models have various predictive powers which needs proper measures to evaluate the classifier. We have used accuracy score and F1-score as the evaluation metrics [8].

*4.2.1 Accuracy Score:* A common metric which is the fraction of the samples correctly predicted. For a predicted value of i[th] sample i.e. $\hat{y}_i$ and $y_i$ being the respective true value, the fraction of right predictions over $n_{\text{samples}}$ may be defined as:

$$\text{accuracy}(y, \hat{y}) = \frac{1}{n_{\text{samples}}} \sum_{i=0}^{n_{\text{samples}}-1} 1(\hat{y}_i = y_i)$$

The mean and standard deviation of the accuracy of the three algorithms has been shown in Table 1.

**Table 1: Mean and deviations of accuracies**

|  | Random Forest | Logistic Regression | SVC |
|---|---|---|---|
| Mean (%) | 71.99 | 66.79 | 64.86 |
| Standard Deviation | 1.55 | 1.39 | 1.46 |

Result for each iteration: Table 3, Figure 8

*4.2.2 F1-Score:* We selected this metric to strike a balance between precision and recall. For $\beta=1$, F1 is derived from:

$$F_\beta = (1 + \beta^2) \frac{\text{precision} \times \text{recall}}{\beta^2 \text{precision} + \text{recall}}.$$

The mean and standard deviation of F1 scores for the three algorithms have been mentioned in Table 2.

**Table 2: Mean and deviations of F1 score**

|  | Random Forest | Logistic Regression | SVC |
|---|---|---|---|
| Mean | 0,70 | 0.63 | 0.58 |
| Standard Deviation | 0.02 | 0.02 | 0.02 |

Result for each iteration: Table 4, Figure 9

### 4.2 Discussion

We started with an unpruned dataset and then ran 20 iterations to prune the dataset to check how the three algorithms performed with each iteration. For the 0[th] iteration, the dataset was unpruned and random forest classifier performed the best as shown in Figure 8 and Figure 9. The accuracy score and F1 score fluctuated as per Table 1 and Table 2 with each iteration, but the ranking of the algorithms remained unchanged.

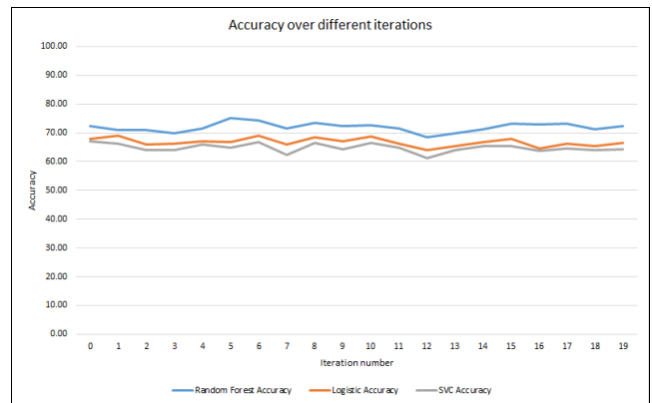

**Figure 8: Accuracy score of each algorithm per iteration**

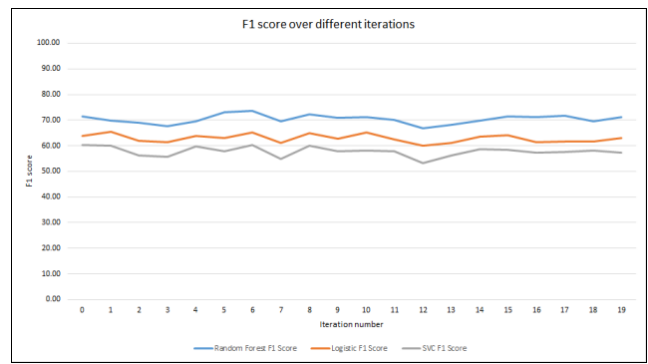

**Figure 9: F1 score of each algorithm per iteration**

### 4.3 Results

Some related works on movie datasets were mostly centered on regression trees while some focused improving SVM accuracy [6] [9]. We ran an unbiased analysis on the three algorithms and observed that random forest performed the best followed by logistic regression and SVC as shown in Figure 8 and Figure 9. Their rankings remain unchanged on unpruned and pruned datasets across the two metrics used. However, several iterations showed some fluctuations in their performance. To conclude, pruning of datasets didn't affect the algorithm performance rankings.

## 5 LIMITATION AND OUTLOOK

The dataset had 5043 data points. The limitation of the dataset was that the classes were not evenly distributed among each class of the target variable as shown in Figure 10. This could result in some class of the data being left out of the train/test set. Future work could include using k-fold cross validation to split the dataset. The work can also be improved by confirming the analysis on a different dataset.

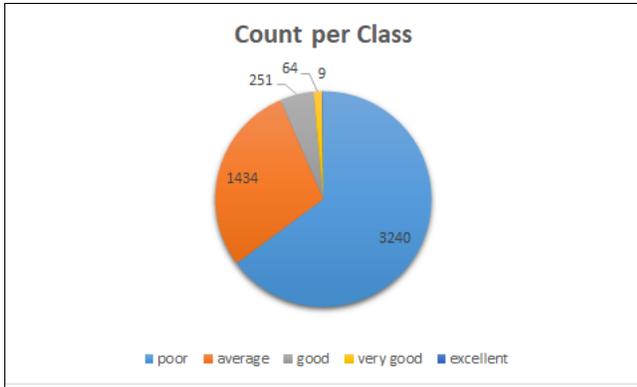

**Figure 10: Distribution of movies in each class**

## 6 ACKNOWLEDGMENT

This work was conducted as part of 2018/19 Machine Learning module CS7CS4/CS4404 at Trinity College Dublin [10].

**Appendix**

**Table 3**: Comparison of Accuracy score of Algorithms (%)

| Random Forest | Logistic Regression | SVC |
|---|---|---|
| 72.50 | 68.00 | 67.00 |
| 71.01 | 68.99 | 66.36 |
| 71.01 | 66.12 | 64.09 |
| 69.81 | 66.33 | 64.07 |
| 71.50 | 67.18 | 66.05 |
| 75.21 | 66.84 | 64.98 |
| 74.35 | 69.16 | 66.77 |
| 71.53 | 65.90 | 62.46 |
| 73.48 | 68.55 | 66.67 |
| 72.42 | 67.05 | 64.42 |
| 72.56 | 68.64 | 66.63 |
| 71.46 | 66.24 | 64.86 |
| 68.48 | 64.10 | 61.22 |
| 69.96 | 65.45 | 64.16 |
| 71.23 | 66.70 | 65.30 |
| 73.24 | 68.04 | 65.55 |
| 72.91 | 64.53 | 63.66 |
| 73.25 | 66.16 | 64.63 |
| 71.38 | 65.35 | 64.04 |
| 72.50 | 66.56 | 64.25 |

**Table 4**: Comparison of F1 score of Algorithms

| Random Forest | Logistic Regression | SVC |
|---|---|---|
| 0.71 | 0.64 | 0.60 |
| 0.70 | 0.65 | 0.60 |
| 0.69 | 0.62 | 0.56 |
| 0.68 | 0.61 | 0.56 |
| 0.70 | 0.64 | 0.60 |
| 0.73 | 0.63 | 0.58 |
| 0.74 | 0.65 | 0.60 |
| 0.70 | 0.61 | 0.55 |
| 0.72 | 0.65 | 0.60 |
| 0.71 | 0.63 | 0.58 |
| 0.71 | 0.65 | 0.58 |
| 0.70 | 0.63 | 0.58 |
| 0.67 | 0.60 | 0.53 |
| 0.68 | 0.61 | 0.56 |
| 0.70 | 0.64 | 0.59 |
| 0.71 | 0.64 | 0.59 |
| 0.71 | 0.61 | 0.57 |
| 0.72 | 0.62 | 0.58 |
| 0.69 | 0.62 | 0.58 |
| 0.71 | 0.63 | 0.57 |